\documentclass[conference]{IEEEtran}
\IEEEoverridecommandlockouts
\usepackage{cite}
\usepackage{amsmath,amssymb,amsfonts}
\usepackage{algorithmic}
\usepackage{graphicx}
\usepackage{textcomp}
\usepackage{hyperref}
\usepackage{xcolor}
\usepackage{array}
\usepackage{multirow}
\usepackage{fancyhdr}
\fancypagestyle{withfooter}{
  
  \fancyhead[L]{}
  \fancyhead[R]{}
  \fancyfoot[C]{\footnotesize Presented at the 2025 IEEE ICRA Workshop on Field Robotics}
}

\def\BibTeX{{\rm B\kern-.05em{\sc i\kern-.025em b}\kern-.08em
    T\kern-.1667em\lower.7ex\hbox{E}\kern-.125emX}}

\begin{document}

\title{Uncertainty Aware Mapping for Vision-Based Underwater Robots\\

\thanks{ $^*$ These authors contributed equally to this work. \\
All the authors are affiliated with the Norwegian University of Science and Technology (NTNU), Trondheim, Norway \\
This material was supported by the Research Council of Norway Awards NO-327292.\\
$\dagger$ These authors are also affiliated with Université de Toulon as part of the Erasmus Mundus Joint Master Degree in Marine and Maritime Intelligent Robotics (EMJMD MIR), received funding from the European Union under the Erasmus+ Programme.
Correspondence: \href{mailto:abhimanb@stud.ntnu.no}{\fontfamily{qcr}\selectfont
 abhimanb@stud.ntnu.no}}
}

\author{ Abhimanyu Bhowmik$^{* \dagger}$, Mohit Singh$^*$, Madhushree Sannigrahi$^\dagger$, Martin Ludvigsen, Kostas Alexis 
}

\maketitle
\thispagestyle{withfooter}
\pagestyle{withfooter}

\begin{abstract}
Vision-based underwater robots can be useful in inspecting and exploring confined spaces where traditional sensors and preplanned paths cannot be followed. Sensor noise and situational change can cause significant uncertainty in environmental representation. Thus, this paper explores how to represent mapping inconsistency in vision-based sensing and incorporate depth estimation confidence into the mapping framework. The scene depth and the confidence are estimated using the RAFT-Stereo model and are integrated into a voxel-based mapping framework, Voxblox. Improvements in the existing Voxblox weight calculation and update mechanism are also proposed. Finally, a qualitative analysis of the proposed method is performed in a confined pool and in a pier in the Trondheim fjord. Experiments using an underwater robot demonstrated the change in uncertainty in the visualization. 

\end{abstract}

% \begin{IEEEkeywords}

% \end{IEEEkeywords}

\section{Introduction} 
\label{sec1}

The oceans and the submerged environments represent some of the least explored regions on Earth. Over the decades, underwater robotic systems, including Remotely Operated Vehicles (ROVs) and Autonomous Underwater Vehicles (AUVs), have played a crucial role in surveying these environments.  However, exploring confined underwater environments requires innovative solutions to ensure safe and efficient navigation. This can include both natural and commercial locations such as shipwrecks, submerged caves, fish farms, ship ballast tanks, etc. 

These areas hold invaluable information, both commercially and scientifically. However, they present exceptional challenges for accurate inspection and exploration. Traditionally, the mapping of underwater bathymetry has been performed using sensors like SONAR, DVL, etc \cite{jacobi2015autonomous}. This mapping is usually done in an open-sea environment where the payload-carrying vehicle follows a preplanned route to map the area. However, a preplanned trajectory is infeasible in confined spaces, as there is a high risk of collision \cite{10791864}. It is important to plan the ROV trajectories online, avoiding collisions and taking into account the dynamic and uncertain nature of the surroundings \cite{Herrmann2024Safe}. The estimation of uncertainty helps not only to improve mapping accuracy but also to develop robust path-planning strategies in cluttered spaces. Thus, this paper focuses on developing uncertainty-aware autonomy for robotic systems, operating in such challenging conditions. Additionally, in small tunnels or tanks, the SONAR image tends to have artifacts near cluttered spaces. These reflections create multi-path returns for the SONAR signal, creating ambiguous or false detections \cite{5191248}. Furthermore, a low-cost, high-resolution inspection for such environments can be better executed using vision-based sensors. Motivated by the above, this paper proposes a vision-based uncertainty-aware mapping strategy for complex underwater environments. 

The structure of the paper is as follows: Section \ref{sec2} contains the proposed methodology, which includes depth and confidence estimation from RAFT-Stereo architecture, a small background Voxblox mapping framework, and how we incorporate confidence into the map. Section \ref{sec3} includes the Experimentation and Results achieved for Confidence estimation and Update. Lastly, Section \ref{sec4} concludes the paper with the future implications of this proposed method.

\begin{figure}
\centerline{\includegraphics[width=0.5\textwidth]{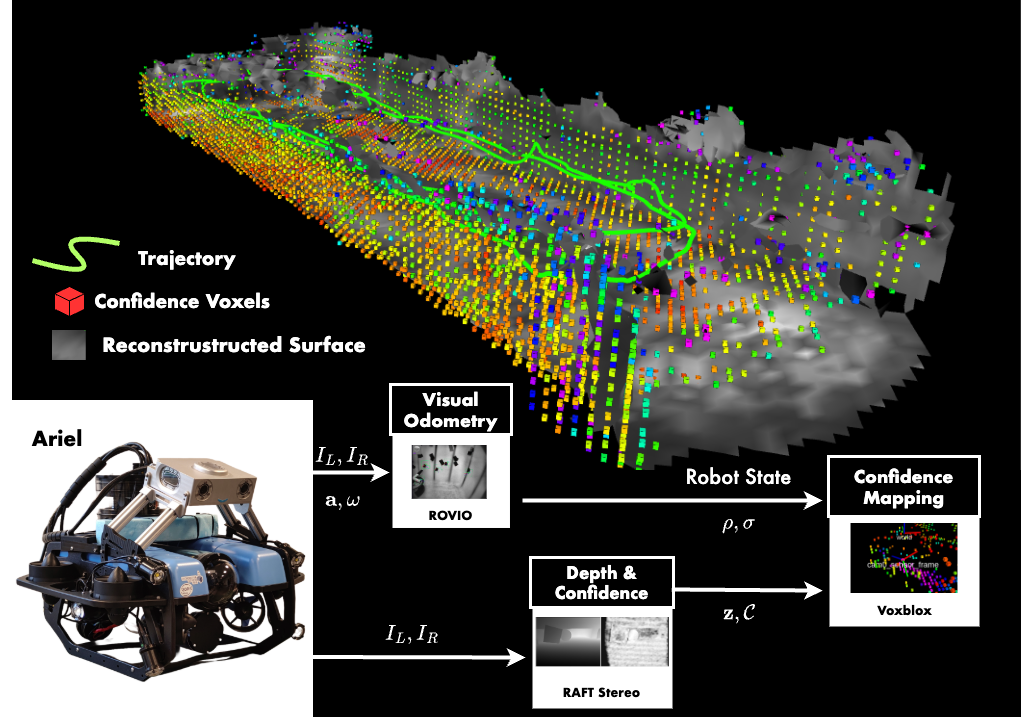}}
\caption{The upper half of the image shows the map of the water tank with 3D surface voxels representing the confidence weights and mesh showing the reconstructed surface. In the bottom left the ROV used for experiments called Ariel is shown with the proposed pipeline. }
\label{pipeline_with_bluerov}
\end{figure}

\section{Methodology} \label{sec2}

In this section, we will discuss our pipeline for including depth confidence in the map. A modified RAFT stereo disparity estimation method is implemented to predict the depth and confidence of a stereo pair of images. We use refractive aquatic robust visual-inertial odometry (ReAqROVIO) \cite{SinghRCMinRovio2024} to estimate the current pose from the camera and IMU sensor data.

\subsection{Stereo depth and confidence estimation}

For our setup, we used a stereo camera to estimate depth using the Recurrent All-Pairs Field Transform (RAFT) framework for optical flow. The RAFT-Stereo \cite{lipson2021raft} extends the principles of the RAFT optical flow method to stereo matching \cite{teed2020raftrecurrentallpairsfield}. This model uses multi-level recurrent updates to refine a high-resolution disparity field iteratively. The choice of RAFT Stereo as a depth estimation method is motivated by two reasons: 
\begin{enumerate}
    \item Low inference time, useful for resource-constrained platforms.
    \item Strong generalizability, making it useful underwater, although it is mostly trained on terrestrial data.
\end{enumerate}

The RAFT Stereo framework has 3 major components: a Feature Encoder to extract a feature vector for every pixel,  a Correlation layer that calculates visual similarity between pixels, and a Recurrent GRU-based update operator that uses correlation volumes to iteratively update the flow field. Given a pair of rectified left and right images($I_L$ and $I_R$), the RAFT Stereo framework aims to find the disparity field ($d$), which represents the horizontal displacements of each pixel in $I_L$ compared to $I_R$.

\begin{figure}
\centering
\includegraphics[width=0.5\textwidth]{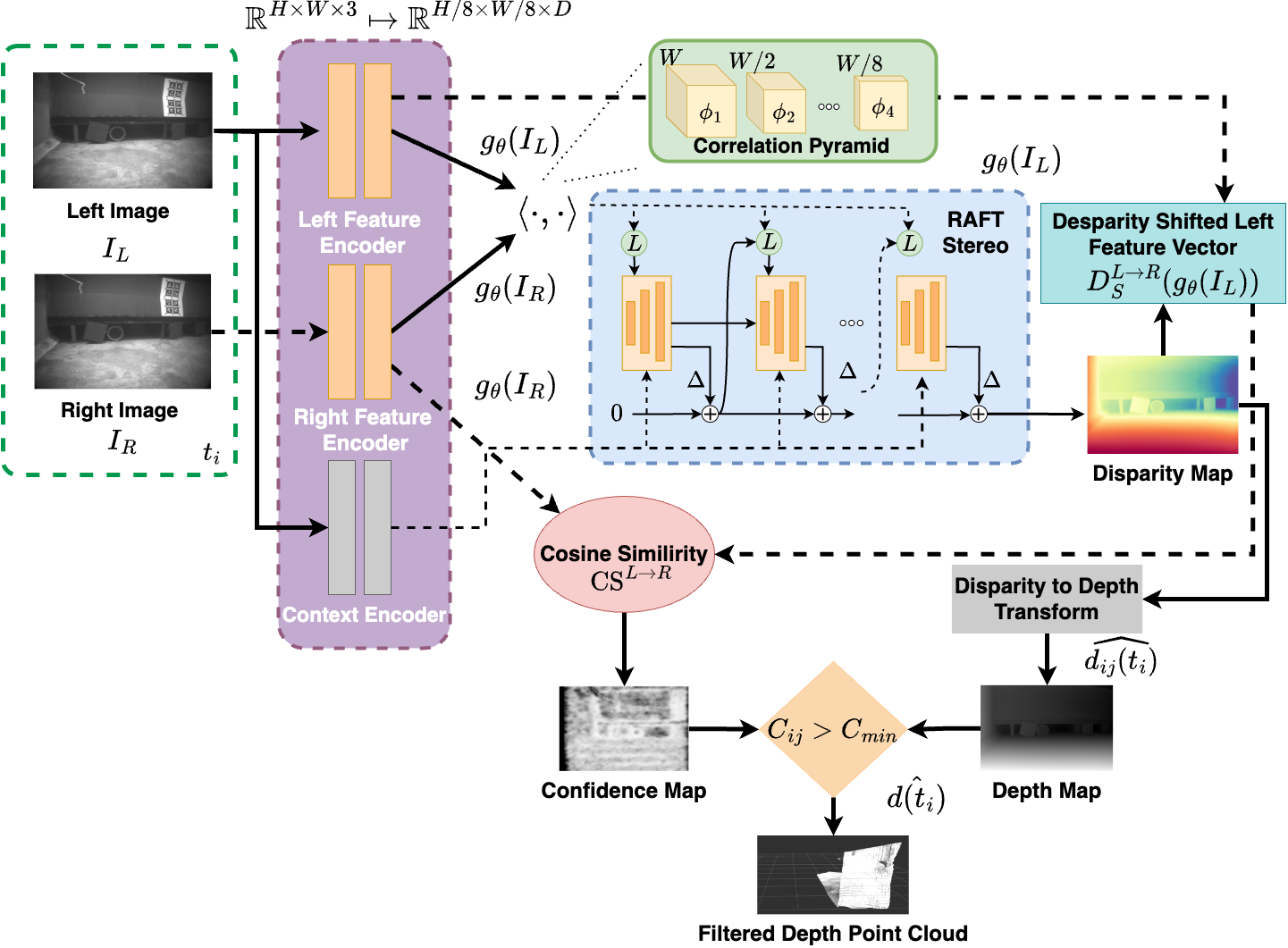}
\caption{Depth and confidence estimation in RAFT Stereo from disparity field. It consists of a feature encoder block followed by a correlation Pyramid by pooling the last 2 dimensions of correlation volume. Then, the RAFT GRU layers complete the stereo disparity estimation pipeline. The disparity map is then transformed into a depth map, and based on the estimated confidence from the similarity between both the initial and transformed feature vector of applicable pixels, the depth point cloud is calculated.}
\label{fig:architecture}
\end{figure}

To estimate depth inconsistency only from the vision-based input, it can be assumed that the left ($I_L$) and right ($I_R$) images contain overlapping features. To measure their visual similarity, we ought to transform the pixels of the first image into another image's frame of reference along the epipolar line. For example, let $g_\theta (\cdot)$ be the feature extractor used in RAFT-Stereo. First, we take both $I_L$ and $I_R$ and calculate their corresponding feature vectors $g_\theta (I_L)$ and $g_\theta (I_R)$. 
For every pixel along the epipolar line, which is present in both $I_L$ and $I_R$, we compute their corresponding transform feature vector $D_S^{L\rightarrow R}(g_\theta (I_L))$. Here, $D_S^{L\rightarrow R}$ is the disparity shift from the left frame to the right frame along the stereo-matching line. 
The depth reconstruction will be more confident if there are substantial overlaps between features of the image pair. The cosine similarity between the corresponding pixel's feature vectors \( g_\theta (I_L) \) and the disparity-shifted feature vectors \( D_S^{L\rightarrow R}(g_\theta (I_L)) \) is computed as:  
\[
\text{CS}^{L\rightarrow R} = \frac{g_\theta(I_L) \boldsymbol{\cdot} D_S^{L\rightarrow R}(g_\theta (I_L))}{\|g_\theta(I_L)\| \ \|D_S^{L\rightarrow R}(g_\theta (I_L))\|}
\]  
Where:   
 \( \boldsymbol{\cdot} \) represents the dot product,  and 
\( \|\cdot\| \) is the \( L_2 \)-norms of the feature vectors.

This cosine similarity produces a confidence map which is used for uncertainty estimation and depth point cloud filtering. Using the stereo-pair geometry, we find the depth map from the disparity using the equation \ref{eq:idk}.

\begin{equation}
Z = \frac{f \cdot B}{d}
\label{eq:idk}
\end{equation}

Where \( Z \) is the depth, \( f \) is the focal length of the camera, \( B \) is the baseline, i.e., the distance between the stereo-camera pair, and \( d \) is the disparity map.

If the confidence exceeds the minimum threshold \(C_{min}\), we take into account the depth map of those pixels, and the rest are discarded. The overall pipeline is described in detail in the figure \ref{fig:architecture}. 

% Again, instead of initializing the disparity field as \(d = 0\) in the multilevel GRU update operator, we initialize it as the reprojection of the disparity field of the previous time step by using \cite{katz2007direct} to remove the occluded regions in the point-cloud. With experimentation, we observed that the latter yields a much better depth representation than the former. The overall pipeline is described in detail in the figure \ref{fig:architecture}.

\subsection{Voxblox: A voxel-based mapping framework}

VoxBlox \cite{oleynikova2017voxblox} is a method for efficiently building volumetric maps using 3D pixel-like blocks called voxels. It incrementally builds Euclidean Signed Distance Fields (ESDFs) from Truncated Signed Distance Fields (TSDFs). ESDFs are spacial representations that store the true Euclidean distance from a point in space to the nearest obstacle or any surface. It encodes both negative distances inside the obstacle and positive distances outside them. TSDF is also a volumetric representation of 3D Space, constructed using point-cloud data. It is essentially the truncated version of ESDF, to a radius around the surface boundary. TSDFs are used to create smooth and noise-resilient surfaces for mapping, which are then converted to ESDFs, which provide distance and gradient information for ROV trajectory optimization and obstacle avoidance. This approach builds TSDFs faster and more accurately than traditional Octomap techniques \cite{wurm2010octomap}. It is more computationally efficient to construct, eliminates sensor noise, and can build human-readable meshes. Instead of computing the Euclidian distance, it uses projective distance, i.e., the distance along the sensor ray to the measured surface, only up to a radius of $\tau$. 

\begin{figure*}
\centerline{\includegraphics[width=0.92\textwidth]{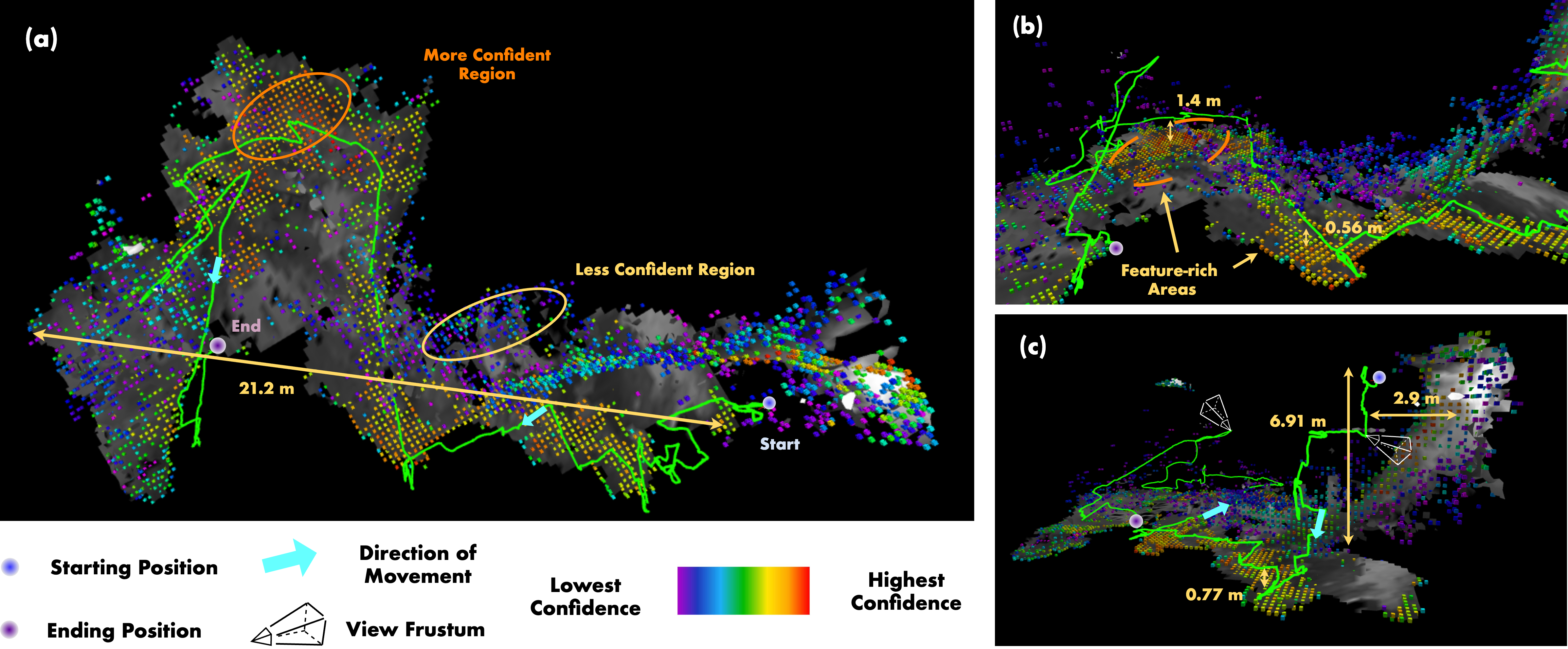}}
\caption{ Visualization of confidence voxels and voxblox's surface reconstruction of our field experiment in a pier in the Trondheim fjord. (a) The top view of the mapped fjord area with start and end positions, showcasing the voxel confidence of different regions.(b) The mapped fjord region displays more confidence in texture-rich areas. (c) The side view of the region with view frustum and direction of motion of ROV during exploration}
\label{fig:result1}
\end{figure*}

For TSDF construction, a typical way to add a new scan is to ray-cast from the sensor's position to each point in the scan data. Then, update the distance and weight values along that line. The merging equations depend on the current weight and distance of a voxel, together with new sensor observations: $\rho$ (distance to the surface) and $\omega$ (weight). Here, $v$ is the voxel's center, $q$ is the position of a 3D point in the sensor data, and $c$ is the sensor's location, with all of them in 3D coordinates $(v, q, c \in \mathbb{R}^3)$. The voxel's updated distance ($\Phi$) and weight ($\Omega$) are calculated as follows:

\begin{equation} \label{eq:distance_calculation}
\rho(v, q, c) = |q - v| \text{sign}((q - v) \cdot (q - c))
\end{equation}
\begin{equation}
\omega_\text{const}(v, q) = 1
\end{equation}
\begin{equation}
\Phi_{k+1}(v, q, c) = \frac{\Omega_k(v)\Phi_k(v) + \omega(v, q)\rho(v, q, c)}{\Omega_k(v) + \omega(v, q)}
\end{equation}
\begin{equation}\label{eq:weight_calculation}
\Omega_{k+1}(v, q) = \min (\Omega_k(v) + \omega(v, q), \Omega_{max})
\end{equation}
The weighting function used here plays an important role since it directly affects how accurate the final 3D reconstruction will be. The proposed weighting function based on a simplified approximation of the RGB-D model is as follows:
\begin{equation} 
\label{eq:weight_function}
\omega_\text{quad}(v, q) = 
\begin{cases}
\frac{1}{z^2} & -\eta < \rho \\
\frac{1}{z^2} \frac{(\rho + \tau)}{\tau - \eta} & -\tau < \rho < -\eta  \\ 0 & \rho < -\tau
\end{cases}
\end{equation}
where $\tau = 4\mu$ is the truncation distance and $\eta = \mu$, and $\mu$ is the voxel size. This essentially means that the weight of the point, if it is inside the $4\mu$ boundary, is proportional to its distance from the voxel center $[\frac{(\rho + \tau)}{\tau - \eta}]$ multiplied by the inverse square from the camera center to the voxels $[\frac{1}{z^2}]$. If the obstacle is outside the $4\mu$ voxel boundary, then the weight is 0. On the other hand, if it lies within the voxel dimension, i.e. $\mu$, then we assume the point and voxel coincide, and it becomes just the inverse square of the voxel distance from the camera center.

\subsection{Incorporate Depth Confidence in the mapping}

The Raft Stereo confidence, which measures how the model is uncertain about the depth at each pixel, is the main estimation of uncertainty we have in the pipeline. There might be other sources of noise that are not taken into account for now, such as variability in state (i.e., the state covariance). However, given a robust VIO method (like ReAqROVIO \cite{singh2023online}), it can be assumed that the Iterated Extended Kalman Filter is consistent if there are no sudden and substantial changes in the environment.

The weight parametrization in TSDF accounts for the nuances of the sensor model and data characteristics. Prior approaches have explored various considerations, such as the angle of incidence between a sensor ray and the surface normal, as demonstrated in KinectFusion \cite{newcombe2011kinectfusion}. Similarly, \cite{nguyen2012modeling} provided an empirical perspective to characterize the sensor noise. Here, the authors cited measurement uncertainty scales with the square of the depth (\( z^2 \)) in the camera frame for Kinect sensors. Voxblox tries to integrate these factors into a unified and adaptive weighting scheme shown in Equation \ref{eq:weight_function}.

To incorporate the confidence of each voxel, we designed a similar formalism to Voxblox. However, the weights are initialized as a function of confidence rather than distance. This is because the uncertainty in stereo depth estimation emerges from the feature-matching error between the stereo-camera setup. Our current strategy for weight initialization is shown below:

% We have compared this strategy with only confidence-based and the voxblox's depth-based strategies and the qualitative results are shown in the table \ref{tab1}.

\begin{equation}
\label{eq:weight_1}
    \omega_{conf} \text(v, q, \mathcal{C}) = \begin{cases}   \mathcal{C}_1 & -\eta < \rho \\   \mathcal{C}_2  & -\tau < \rho < -\eta \\ 0 & \rho < -\tau \end{cases}
\end{equation}
Where, $\mathcal{C}_1$ is the confidence of the pixel at the voxel's location and $\mathcal{C}_2$ is the confidence of the edge (closest obstacle) pixel. We have compared this with other strategies where weight is both the function of confidence and depth similar to Voxblox. However, the distance from the surface has a predominant impact on initialized weights in that case, and the effect of feature-based confidence was not realized. This is not necessarily true since a texture-rich surface will have a low matching uncertainty and thus possess more confidence than a texture-less surface, even if the latter is closer to the camera. 

Now, to obtain the correspondence of the 3D point (Voxel or Edge Pixel) in the 2D confidence plane, we use the camera matrix. According to the general formulation of the camera matrix:
\begin{equation}
\begin{bmatrix}
a \\
b \\
1
\end{bmatrix}
=\begin{bmatrix}
\alpha_x & 0 & o_x \\
0 & \alpha_y & o_y \\
0 & 0 & 1
\end{bmatrix}
\begin{bmatrix}
X/Z \\
Y/Z \\
1
\end{bmatrix}\\
\end{equation}
Where $\alpha_x$ and $\alpha_y$ are the focal lengths and $o_x$ and $o_y$ are the center coordinates of the confidence frame (coinciding with the image frame). $X$, $Y$ and $Z$ is the location of the corresponding 3D point coordinate.

The weight update process is quite similar to the original weight update rule in Voxblox, as it balances performance and efficiency. However, the equation \ref{eq:weight_calculation} accumulates the weights instead of updating them. This suggests that even if the confidence is not increasing, the weight keeps accumulating with time. To counter this, the equation has been modified to \ref{eq:weight_calculation_new}, which updates the weight with the average of current and previous weights. Note that one could also use a moving average instead of a simple average, although it will increase the computational overhead without any major improvement in accuracy. 

\begin{equation}
\label{eq:distance_calculation_new}
  \rho(v, q, c) = \|q - v\| \text{sign}((q - v) \cdot (q - c))  
\end{equation}
\begin{equation}
\label{eq:weight_2}
    \omega = \omega_{conf} \text(v, q, \mathcal{C})
\end{equation}
\begin{equation}
   \Phi_{k+1}(v, q, c) = \frac{\Omega_k(v)\Phi_k(v) + \omega(v, q)\rho(v, q, c)}{\Omega_k(v) + \omega(v, q)} 
\end{equation}
\begin{equation}\label{eq:weight_calculation_new}
   \Omega_{k+1}(v, q) = \min \left[ \frac{\Omega_k(v) + \omega(v, q)}{2}, \Omega_{max}\right] 
\end{equation}

\section{Experiments and Results}
\label{sec3}

A custom ROV called Ariel, built on top of the BlueROV2 Heavy Configuration as shown in figure \ref{pipeline_with_bluerov} was used for experiments.
It was manually piloted to perform varying trajectories while collecting synchronized stereo camera frames and IMU data. The system integrates an Alphasense Core Research Kit with five monochrome Sony IMX-287 global-shutter cameras (0.4MP, FOV in air: $165.4^\circ \times 126^\circ \times 92.4^\circ$, 2.4mm focal length), synchronized with a Bosch BMI085 IMU. Data is transmitted via Gigabit Ethernet to an onboard NVIDIA Orin compute board. Both units are waterproofed and mounted, with the Alphasense tilted 16° downward.

The experiments were performed in  2 different environments: an enclosed lab setting and an outdoor field trial. The controlled indoor experiments were carried out at the NTNU Marine Cybernetics Laboratory (MC-lab), which has a water tank of dimension $40\text{m} \times 6.45\text{m} \times 1.5\text{m}$. The ROV was operated to explore more than half of the total pool length, as in figure \ref{fig:result3}. It started from the blue starting point and completed 2 laps in the pool. A similar open-source dataset published by Autonomous Robots Lab can be found online \cite{singh2023online}. The field experiments were conducted in a pier in the Trondheim fjord at about $7\text{m}$ depth, as seen in figure \ref{fig:result1}.

\subsection{Depth and Confidence Estimation}

\begin{figure}
\centering
\includegraphics[width=0.5\textwidth]{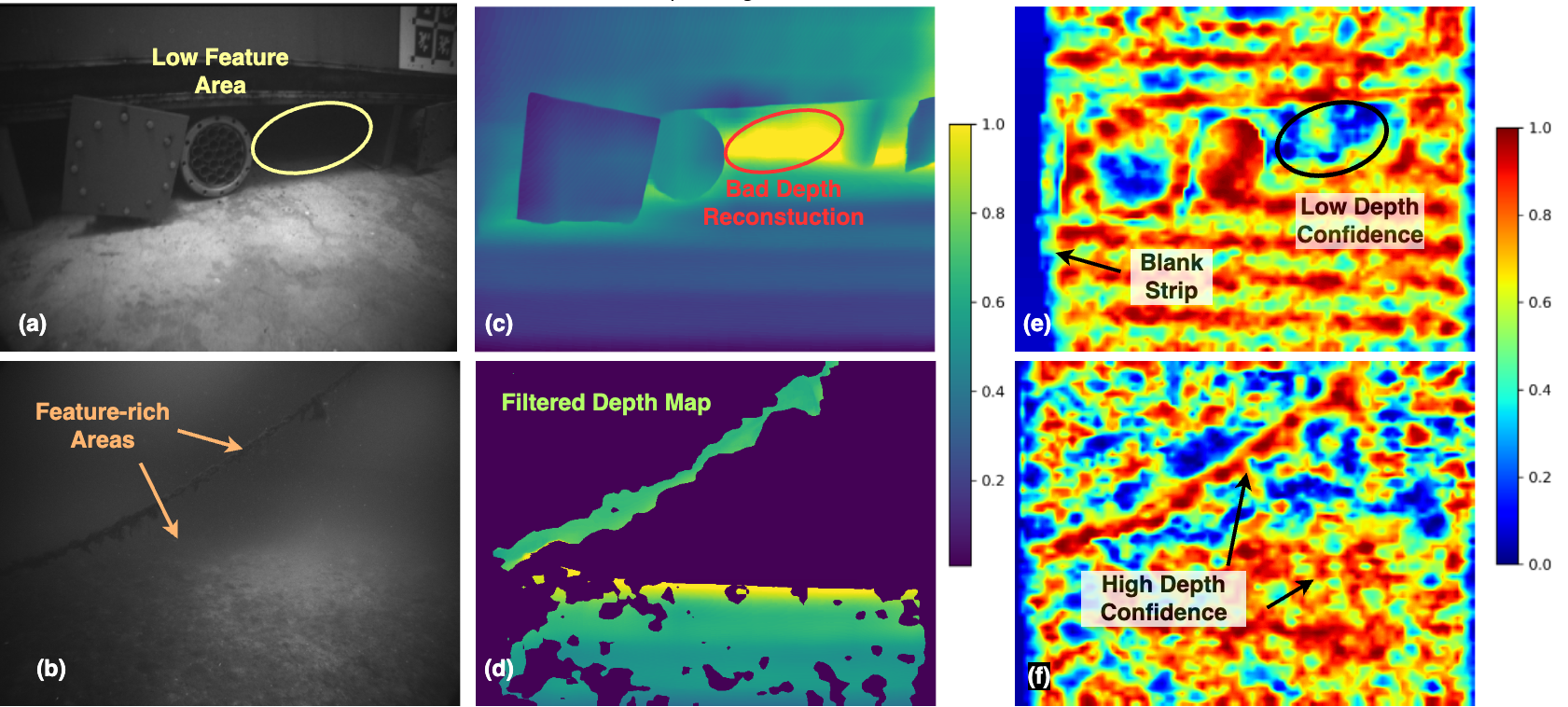}
\caption{Visualization of image, depth map and confidence map. (a) and (b) Image from the left stereo camera in the pool and fjord, (c) and (d) Normalized Estimated depth from RAFT Stereo pipeline, (e) and (f) Normalized Estimated depth confidence for respective pixels }
\label{fig:results2}
\end{figure}

The results of the depth estimation and its confidence are consistent with the general expectations, as noticed in figure \ref{fig:results2}. In the figure, (a) and (b) represent the indoor and outdoor conditions, respectively. The water in the controlled tank has higher visibility due to a well-lit environment and a lower depth profile. However, the fjord water has a higher turbidity, resulting in reduced visibility, which negatively affects feature detection.

The depth reconstruction depends on the network's ability to match features. Given a scene with rich features and contrasts, the probability of getting a good stereo depth estimation is higher. This contrast can be visualized by comparing the reconstruction of the floors of both environments. The fjord bed is texture-rich, which results in higher depth confidence. On the other hand, the marked area on the pool floor is dark and featureless and has a bad predicted depth, consequently resulting in lower confidence. Depth points with low confidence have been filtered out if their values fall below the minimum threshold. This resulted in blank patches in the depth map [Fig. \ref{fig:results2} (d)].

Another observation is the feature-vector transformation effect. In the confidence plane, there is a noticeable skew towards the left. This is likely due to the transformation of the left feature vector into the right feature vector while calculating the cosine similarity among them. Such a shift to the right resulted in a wider blank strip towards the left, shown in Fig. \ref{fig:results2} (e). The structure of this blank strip varies frame by frame since it depends on the transformation of pixel locations of prominent features in the image. 

% This blank strip is not caused by the vignette effect but rather by the transformation of pixel locations during the matching process.

% \begin{table}
% \centering
% \caption{Analysing different strategies for weight initilization}
% \label{tab1}
% \begin{tabular}{ | c | c| c | }
% \hline
% \textbf{Weight Initlization} & \textbf{Weight Dependency}       & \textbf{Appearance}             \\ \hline
% \textbf{ \(\frac{1}{z^2} \)}     & \text{Inverse depth only}        & \text{Smooth gradient}          \\ \hline
% \textbf{\(\mathcal{C}\)}     & \text{Confidence only}           & \text{Patchy, discrete}         \\ \hline
% \textbf{\(\frac{1}{z^2} \cdot \mathcal{C}\)}     & {Confidence + Inverse depth} & {Smooth + confidence-aware} \\ 
% \hline
% \end{tabular}
% \end{table}

\subsection{Confidence update and Visualization}

The odometry of the ROV is estimated using the ReAqROVIO framework, as mentioned above. The surface along the trajectory is reconstructed using Voxblox mesh rendering, depicted in grayscale in figure \ref{pipeline_with_bluerov}. Each voxel depicts the depth confidence of the pool surface using the VIBGYOR color scheme, where Violet indicates the lowest confidence and Red signifies the highest. This means that, given sufficiently accurate odometry, the system has a more confident reconstruction of the areas where the voxels are primarily red. 

As we observe in figure \ref{fig:result1}, most of the features in the fjord are present on the subaqueous fjordbed. Areas marked as relatively more confident are the ones that have a texture-rich surface and occupy a significant part of the viewing frustum. The depth confidence appears higher when closer to the surface since the features are less visible with increasing distance. However, in good visibility, the confidence appears to be more dependent on the textures. As portrayed in figure \ref{fig:result1} (b), two texture-rich regions at different depths exhibit similar confidence. Although the surfaces are at 1.4m and 0.56m apart from the ROV, the former region is more confident due to distinctly visible features.

\begin{figure}[!t]
\centering\raggedbottom
\includegraphics[width=0.45\textwidth]{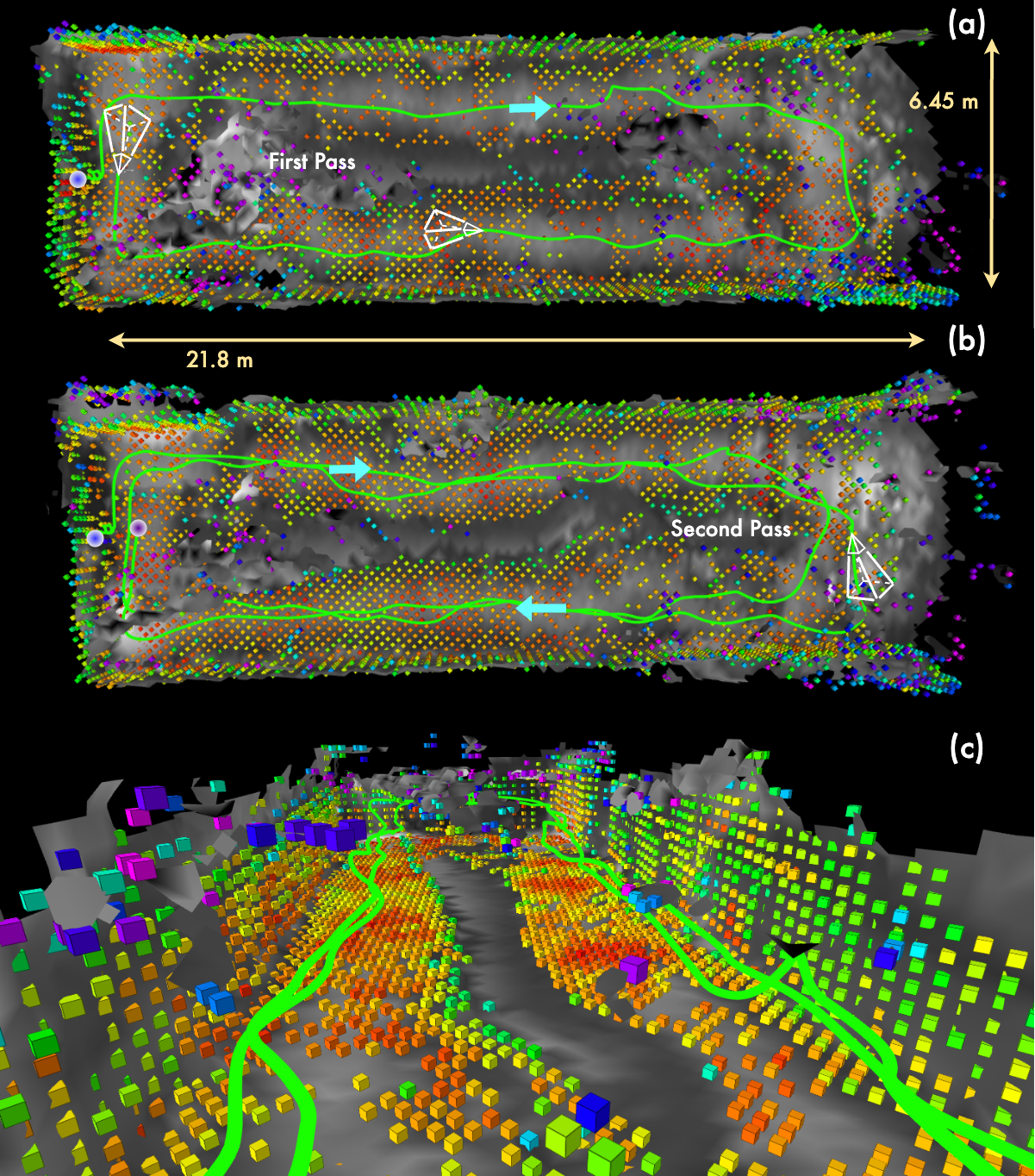}
\caption{Mapping in the indoor pool environment at MC-Lab. (a) shows the top view after a single lap, (b) shows the top view with 2 lap, and (c) gives a close-up lateral view of the map}
\label{fig:result3}
\end{figure}

The result for an indoor, well-lit environment is shown in figure \ref{fig:result3}. The ROV follows a similar trajectory twice in the pool and maps the area as demonstrated in fig \ref{fig:result3} (a) and (b). We notice that the system is sufficiently reliable with one lap, and the confidence map generated is similar to that of two laps. Additionally, for the surfaces that are not in front of the frustum, i.e., the ROV is moving parallel to the surface (and not facing it directly), the confidence is significantly lower. In the figure \ref{fig:result3} (c), we observe that the confidence is higher in the texture-rich tank floor as the camera faces the surface directly, while the pool walls show relatively less confidence. 

\vspace{-3pt}
\section{Conclusion} 
\label{sec4}

In this paper, we proposed a methodology for depth confidence estimation and incorporated it into a TSDF-based 3D mapping framework. We presented the dependence of depth confidence over the surface textures. We also demonstrated that revisiting a place doesn't guarantee confidence increment unless the view frustum is changing significantly at each lap. Additionally, we found that in field experiments, being closer to a surface indirectly contributes to visibility in underwater environments, affecting the confidence map. Incorporating depth and other phenomena, such as optical attenuation and turbidity, in a regularised manner could be a potential future possibility for this work. Likewise, the proposed formulation could be extended to exploratory planning as well. It will enable the planner to not only focus on efficient exploration of regions but also maximize the mapping confidence of previously explored areas.

\vspace{-4pt}

\section{Disclaimer}

 The European Commission's support for the production of this publication does not constitute an endorsement of the contents, which reflect the views only of the authors, and the Commission cannot be held responsible for any use that may be made of the information contained therein.
\vspace{-2pt}
% \section*{References}
\bibliographystyle{IEEEtran}
\bibliography{bibliography.bib}

\begin{thebibliography}{10}
\providecommand{\url}[1]{#1}
\csname url@samestyle\endcsname
\providecommand{\newblock}{\relax}
\providecommand{\bibinfo}[2]{#2}
\providecommand{\BIBentrySTDinterwordspacing}{\spaceskip=0pt\relax}
\providecommand{\BIBentryALTinterwordstretchfactor}{4}
\providecommand{\BIBentryALTinterwordspacing}{\spaceskip=\fontdimen2\font plus
\BIBentryALTinterwordstretchfactor\fontdimen3\font minus \fontdimen4\font\relax}
\providecommand{\BIBforeignlanguage}[2]{{%
\expandafter\ifx\csname l@#1\endcsname\relax
\typeout{** WARNING: IEEEtran.bst: No hyphenation pattern has been}%
\typeout{** loaded for the language `#1'. Using the pattern for}%
\typeout{** the default language instead.}%
\else
\language=\csname l@#1\endcsname
\fi
#2}}
\providecommand{\BIBdecl}{\relax}
\BIBdecl

\bibitem{jacobi2015autonomous}
M.~Jacobi, ``Autonomous inspection of underwater structures,'' \emph{Robotics and Autonomous Systems}, vol.~67, pp. 80--86, 2015.

\bibitem{10791864}
X.~Wang, Z.~Sha, and F.~Zhang, ``Adaptive integral sliding mode control for attitude tracking of underwater robots with large range pitch variations in confined spaces,'' \emph{IEEE Robotics and Automation Letters}, vol.~10, no.~2, pp. 979--986, 2025.

\bibitem{Herrmann2024Safe}
F.~Herrmann, S.~Zach, J.~Banfi, J.~Peters, G.~Chalvatzaki, and D.~Tateo, ``Safe and efficient path planning under uncertainty via deep collision probability fields,'' \emph{IEEE Robotics and Automation Letters}, vol.~9, pp. 9327--9334, 2024.

\bibitem{5191248}
B.~J. Davis, P.~T. Gough, and B.~R. Hunt, ``Modeling surface multipath effects in synthetic aperture sonar,'' \emph{IEEE Journal of Oceanic Engineering}, vol.~34, no.~3, pp. 239--249, 2009.

\bibitem{SinghRCMinRovio2024}
M.~Singh and K.~Alexis, ``Online refractive camera model calibration in visual inertial odometry,'' in \emph{2024 IEEE/RSJ International Conference on Intelligent Robots and Systems (IROS)}, 2024, pp. 12\,609--12\,616.

\bibitem{lipson2021raft}
L.~Lipson, Z.~Teed, and J.~Deng, ``Raft-stereo: Multilevel recurrent field transforms for stereo matching,'' in \emph{International Conference on 3D Vision (3DV)}, 2021.

\bibitem{teed2020raftrecurrentallpairsfield}
\BIBentryALTinterwordspacing
Z.~Teed and J.~Deng, ``Raft: Recurrent all-pairs field transforms for optical flow,'' 2020. [Online]. Available: \url{https://arxiv.org/abs/2003.12039}
\BIBentrySTDinterwordspacing

\bibitem{oleynikova2017voxblox}
H.~Oleynikova, Z.~Taylor, M.~Fehr, R.~Siegwart, and J.~Nieto, ``Voxblox: Incremental 3d euclidean signed distance fields for on-board mav planning,'' in \emph{IEEE/RSJ International Conference on Intelligent Robots and Systems (IROS)}, 2017.

\bibitem{wurm2010octomap}
K.~M. Wurm, A.~Hornung, M.~Bennewitz, C.~Stachniss, and W.~Burgard, ``Octomap: A probabilistic, flexible, and compact 3d map representation for robotic systems,'' in \emph{Proc. of the ICRA 2010 workshop on best practice in 3D perception and modeling for mobile manipulation}, vol.~2, 2010, p.~3.

\bibitem{singh2023online}
M.~Singh, M.~Dharmadhikari, and K.~Alexis, ``An online self-calibrating refractive camera model with application to underwater odometry,'' 2023.

\bibitem{newcombe2011kinectfusion}
R.~A. Newcombe, S.~Izadi, O.~Hilliges, D.~Molyneaux, D.~Kim, A.~J. Davison, P.~Kohi, J.~Shotton, S.~Hodges, and A.~Fitzgibbon, ``Kinectfusion: Real-time dense surface mapping and tracking,'' in \emph{2011 10th IEEE international symposium on mixed and augmented reality}.\hskip 1em plus 0.5em minus 0.4em\relax Ieee, 2011, pp. 127--136.

\bibitem{nguyen2012modeling}
C.~V. Nguyen, S.~Izadi, and D.~Lovell, ``Modeling kinect sensor noise for improved 3d reconstruction and tracking,'' in \emph{2012 second international conference on 3D imaging, modeling, processing, visualization \& transmission}.\hskip 1em plus 0.5em minus 0.4em\relax IEEE, 2012, pp. 524--530.

\end{thebibliography}

\end{document}